\documentclass[final,5p,times,twocolumn,nonatbib]{elsarticle}
\usepackage[T1]{fontenc}
\usepackage[american]{babel}
\usepackage{amsmath,amssymb,amsfonts}
\usepackage{graphicx}
\usepackage{booktabs,multirow,array,adjustbox}
\usepackage{caption}
\usepackage{needspace}
\usepackage{placeins}
\usepackage{float}
\usepackage{textcomp}
\usepackage{csquotes}
\usepackage[backend=biber,style=apa]{biblatex}
\usepackage{xurl}
\usepackage[hidelinks]{hyperref}
\newcommand{\cmark}{\checkmark}
\newcommand{\xmark}{\times}
\journal{Neural Networks}

\begin{document}
\begin{frontmatter}
\title{CHRep: Cross-modal Histology Representation and Post-hoc Calibration for Spatial Gene Expression Prediction}
\author[bupt,lab]{Changfan Wang}
\ead{2025018043@bupt.cn}
\author[hospital]{Xinran Wang}
\ead{wxrsarah924@163.com}
\author[hospital]{Donghai Liu}
\ead{345818520@qq.com}
\author[bupt,lab]{Fei Su}
\ead{sufei@bupt.edu.cn}
\author[hospital]{Lulu Sun}
\ead{lulusun@bjmu.edu.cn}
\author[bupt,lab]{Zhicheng Zhao\corref{cor1}}
\ead{zhaozc@bupt.edu.cn}
\author[bupt,lab]{Zhu Meng\corref{cor1}}
\ead{bamboo@bupt.edu.cn}
\cortext[cor1]{Corresponding authors.}
\affiliation[bupt]{organization={Beijing University of Posts and Telecommunications},
  city={Beijing},postcode={100876},country={China}}
\affiliation[lab]{organization={Beijing Key Laboratory of Network System and Network Culture},
  city={Beijing},postcode={100876},country={China}}
\affiliation[hospital]{organization={Peking University Third Hospital},
  city={Beijing},postcode={100191},country={China}}

\begin{abstract}
Spatial transcriptomics (ST) enables spatially resolved gene profiling but remains expensive and low-throughput, limiting large-cohort studies and routine clinical use. Predicting spatial gene expression from hematoxylin and eosin (H\&E) slides provides a computational alternative, with slide-level appearance variation and regression-driven over-smoothing posing key challenges. CHRep is introduced as a two-stage framework combining structure-aware representation learning with post-hoc calibration. The representation-learning stage jointly optimizes correlation-aware regression, symmetric image--expression alignment, and coordinate-induced spatial topology regularization. The calibration stage combines a similarity-weighted estimate from a training gallery with a magnitude-regularized residual predictor operating on frozen histology features. This design integrates spatially informed cross-modal learning with retrieval-based prediction refinement. Under leave-one-slide-out evaluation on three cohorts, CHRep achieves the highest Pearson correlation coefficients on all considered genes [PCC(ACG)] and the top-50 highly expressed genes among the compared methods. PCC(ACG) reaches 0.2733 on HER2+, 0.3532 on cSCC, and 0.2718 on Alex+10x. CHRep also achieves the lowest mean squared error (MSE) and mean absolute error (MAE) on cSCC (0.3983 and 0.4048) and Alex+10x (0.2102 and 0.3545). The official CHRep repository is available at \url{https://github.com/changfan-wang/CHRep}.
\end{abstract}
\end{frontmatter}

\section{Introduction}
Spatial transcriptomics (ST) has rapidly advanced as a key technology for interrogating tissue biology with spatially resolved molecular readouts \parencite{alon2021expansion,chen2022spatiotemporal}. By measuring genome-wide RNA abundance while retaining the spatial coordinates of capture locations, ST enables the study of spatial gene expression programs \emph{in situ}, and supports analyses that are difficult for bulk assays, such as identifying spatial domains, characterizing microenvironmental niches, and probing cell--cell communication patterns \parencite{asp2020spatially,rao2021exploring,hu2023deciphering}. To better exploit these attributes, a broad range of computational methods has been developed for spatial representation learning and domain discovery \parencite{ren2022identifying,LI2026109155}, super-resolution mapping \parencite{bergenstrahle2022xfuse,zhang2024istar}, and denoising/imputation of sparse measurements, aiming to recover coherent and fine-grained spatial expression landscapes \parencite{zhao2021spatial}.

In practice, deploying ST at scale remains constrained by experimental cost, limited throughput, and platform-dependent noise such as dropouts and batch artifacts \parencite{lee2022recent}. In contrast, routine H\&E whole-slide images are inexpensive, standardized in clinical workflows, and available at large scale across cohorts. This contrast has motivated a rapidly growing direction that seeks to predict spatial gene expression from histopathology morphology, approximating ST-like molecular readouts when only slides are available and enabling large-cohort studies without additional wet-lab assays \parencite{he2020integrating,xie2023spatially,min2024multimodal,monjo2022deepspace}.

However, inferring transcriptomic states from morphology is fundamentally challenging. First, histology is an indirect surrogate where similar visual patterns may correspond to distinct transcriptional programs \parencite{pizurica2024digital}. Second, each spot aggregates heterogeneous microenvironments, making spatially localized expression patterns difficult to recover. Objectives dominated by pointwise regression can favor conservative predictions that attenuate gene-wise variation and spatial expression contrast. Finally, realistic deployment demands robustness against slide-specific staining variations, which induce distribution shifts that degrade generalization without explicit alignment mechanisms \parencite{tellez2019quantifying,LIU2026108805}.

Representative histology-to-ST predictors can be broadly grouped by the inductive biases they emphasize for bridging morphology and molecular profiles \parencite{Yang_2023_WACV,chung2024accurate}. Early regression-based pipelines learn patch-to-gene mapping directly from paired H\&E--ST data, exemplified by ST-Net \parencite{he2020integrating}, while transformer-style contextual modeling, such as HisToGene, captures broader histological dependencies \parencite{pang2021leveraging}. To further exploit spatial context, methods including Hist2ST and THItoGene incorporate structured aggregation across spots using graph/transformer or hybrid modules \parencite{zeng2022spatial,jia2023thitogene}. More recently, multimodal embedding approaches seek a shared image--expression space to improve correspondence and transferability \parencite{yan2025magic}: BLEEP performs retrieval-based expression transfer via cross-modal similarity, whereas mclSTExp combines spatial encoding with contrastive objectives for representation alignment \parencite{xie2023spatially,min2024multimodal}.

Despite this progress, practical challenges remain under realistic leave-one-slide-out evaluation. Slide-level appearance and protocol variations can perturb visual features, weakening histology--expression correspondence and leading to brittle generalization \parencite{stacke2020measuring}. Moreover, objectives dominated by pointwise regression can encourage conservative predictions that attenuate spatial variation and reduce gene-wise correlation, even when absolute errors appear competitive. Finally, while many designs enhance local context aggregation, they often lack explicit constraints to preserve higher-order spatial relationships beyond immediate neighbors, which can gradually dilute mesoscopic tissue organization in the learned embedding space.

To address these challenges, CHRep combines structure-aware representation learning with post-hoc calibration. The first stage integrates correlation-aware regression, symmetric image--expression alignment, and coordinate-induced multi-hop topology regularization. The second stage refines predictions through similarity-weighted neighbor estimation and magnitude-regularized residual correction on frozen histology features.

The main contributions are summarized as follows:
\begin{itemize}
    \item A unified representation-learning framework is developed by integrating correlation-aware regression, symmetric image--expression alignment, and coordinate-induced multi-hop topology regularization for spatial gene expression prediction.

    \item A post-hoc calibration scheme combines similarity-weighted neighbor estimation with magnitude-regularized residual correction, refining expression predictions on frozen histology features.

    \item Experiments on cSCC~\parencite{ji2020multimodal}, HER2+~\parencite{andersson2020spatial}, and Alex+10x~\parencite{wu2021single} demonstrate the highest PCC(ACG) and PCC(HEG@50) among the compared methods on all three datasets under leave-one-slide-out evaluation, together with the lowest MSE and MAE on cSCC and Alex+10x.
\end{itemize}

\Needspace{3\baselineskip}

\section{Methodology}

\begin{figure*}[t]
    \centering
    \includegraphics[width=\textwidth]{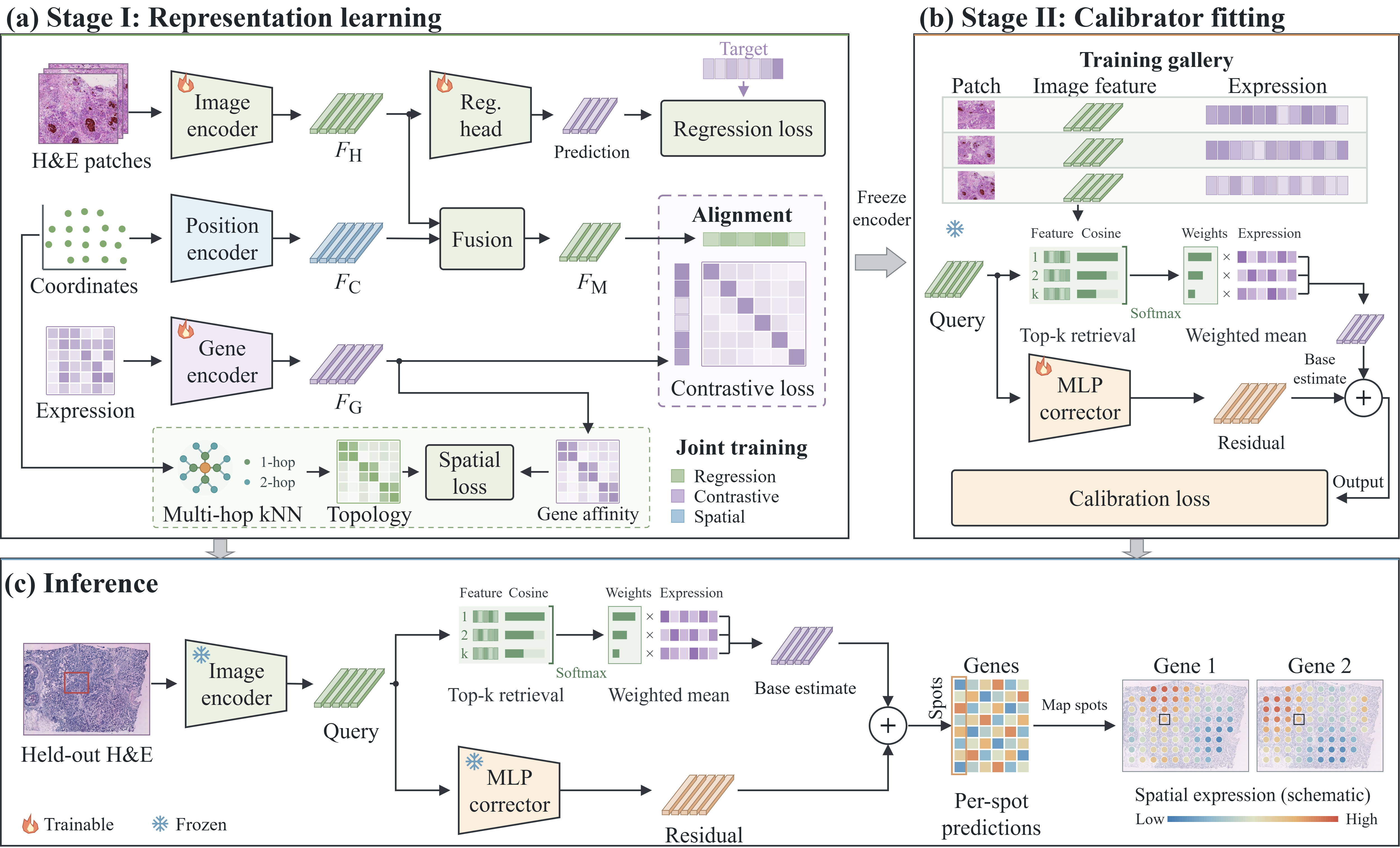} 

    \caption{Overview of CHRep. \textbf{(a) Stage I: representation learning.} Histology features $F_H$ support expression regression and are fused with coordinate features $F_C$ to form $F_M$ for cross-modal alignment with gene features $F_G$. Coordinate-induced multi-hop topology regularizes the similarity structure of $F_G$. \textbf{(b) Stage II: calibrator fitting.} The image encoder is frozen, and a residual predictor is trained to refine a similarity-weighted expression estimate from the training gallery. \textbf{(c) Inference.} The fixed image encoder and residual predictor generate spot-level expression predictions by combining the retrieved estimate with the learned correction.}
\label{fig:method_overview}

\end{figure*}

\subsection{Overall Framework}
Fig.~\ref{fig:method_overview} presents the two stages of CHRep: representation learning and post-hoc calibration. During representation learning, each spot is represented by an H\&E patch, its spatial coordinate, and its gene expression vector. The histology branch supports expression regression, the coordinate branch provides position-aware features for cross-modal alignment, and the gene branch encodes expression profiles. These branches are jointly optimized using correlation-aware regression, symmetric cross-modal alignment, and coordinate-induced topology regularization.

During post-hoc calibration, the image encoder is frozen and the Correction Module is fitted on the training slides. The Estimate Module aggregates expression profiles from similar training-gallery embeddings, while the Correction Module predicts an additive residual from the query feature. At inference, the fixed encoder and calibration module produce the final prediction by summing the estimated expression and residual correction.
\subsection{Data Preparation and Notation}

For each tissue section, each spatial location (spot) is represented by a triplet $(\mathbf{x}, \mathbf{g}, \mathbf{p})$, where $\mathbf{x}\in\mathbb{R}^{3\times H \times W}$ denotes an H\&E image patch centered at the spot, $\mathbf{g}\in\mathbb{R}^{G}$ denotes the corresponding spot-level expression vector restricted to a fixed highly-variable gene (HVG) set, and $\mathbf{p}\in\mathbb{R}^{2}$ denotes the 2D spot coordinate. Raw counts are preprocessed by library-size normalization followed by log transformation, and the HVG list is fixed for each cohort to ensure consistency in training and evaluation.

Within each leave-one-slide-out fold, expression is further standardized on a per-gene basis using the mean $\boldsymbol{\mu}$ and standard deviation $\boldsymbol{\sigma}$ computed from the training slides only, yielding
$\tilde{\mathbf{g}}=(\mathbf{g}-\boldsymbol{\mu})\oslash(\boldsymbol{\sigma}+\epsilon)$.
Representation learning, gallery-based estimation, and evaluation use the standardized expression space $\tilde{\mathbf{g}}$, placing each gene on a common training-derived scale.

Fixed-size image patches are cropped around each spot center in pixel space. For cohorts with pixel-aligned spot centers, patch extraction is performed directly on the whole-slide image using the recorded coordinates. For $10\times$ Visium data, spot coordinates are first transformed into the image coordinate system using the provided scale factor, and boundary-safe clamping is then applied to spots near slide borders prior to cropping. During training, lightweight data augmentations are employed, including color jitter, random horizontal flips, and random rotations. During inference, deterministic preprocessing is adopted without augmentation.

Spatial coordinates guide representation learning through two paths: coordinate features $F_C$ are fused with histology features $F_H$ for cross-modal alignment, and coordinate-induced neighborhoods define the topology target for gene features $F_G$.

\subsection{Representation Learning with a Unified Objective}
\label{sec:rep_learning}

During Stage I, CHRep learns from mini-batches of paired H\&E patches, spatial coordinates, and gene expression vectors, denoted by $\{(\mathbf{x}_i,\mathbf{p}_i,\mathbf{g}_i)\}_{i=1}^{B}$. The image, coordinate, and gene encoders produce histology features $F_H$, coordinate features $F_C$, and expression features $F_G$, respectively. Histology and coordinate features are fused into $F_M$ for cross-modal alignment.
For each spot $i$, the image encoder $f_{\theta}$ extracts the histology feature
\begin{equation}
F_H(i)=f_{\theta}(\mathbf{x}_i),
\label{eq:fh}
\end{equation}
and the regression head predicts the standardized expression vector from the image feature:
\begin{equation}
\hat{\mathbf{g}}_i=h_{\phi}\!\left(F_H(i)\right).
\label{eq:enc_reg}
\end{equation}

In parallel, the coordinate branch maps the 2-D spot coordinate $\mathbf{p}_i=(p_i^x,p_i^y)$ to a position-aware feature,
\begin{equation}
F_C(i)=e_{\psi}(\mathbf{p}_i).
\label{eq:fc}
\end{equation}
The resulting coordinate feature is fused with the histology feature to form the coordinate-guided morphology representation $F_M(i)$ used for subsequent cross-modal alignment. Meanwhile, the gene branch maps the standardized expression vector to a gene representation,
\begin{equation}
F_G(i)=g_{\omega}\!\left(\tilde{\mathbf{g}}_i\right).
\label{eq:fg}
\end{equation}

The regression branch is supervised by a correlation-aware objective,
\begin{equation}
\begin{aligned}
\mathcal{L}_{\mathrm{reg}}
&=\frac{1}{BG}\sum_{i=1}^{B}\left\|\hat{\mathbf{g}}_i-\tilde{\mathbf{g}}_i\right\|_2^2
+\lambda_{\mathrm{mae}}\frac{1}{BG}\sum_{i=1}^{B}\left\|\hat{\mathbf{g}}_i-\tilde{\mathbf{g}}_i\right\|_1 \\
&\quad +\lambda_{\mathrm{PCC}}\left(1-\overline{\mathrm{PCC}}\right),
\end{aligned}
\label{eq:lreg}
\end{equation}
where $\overline{\mathrm{PCC}}$ denotes the average gene-wise Pearson correlation computed across the mini-batch.

Lightweight projection heads map the coordinate-guided morphology features $F_M$ and gene features $F_G$ into normalized embeddings for cross-modal alignment. A symmetric contrastive objective \parencite{radford2021learning} is defined as
\begin{equation}
\mathcal{L}_{\mathrm{con}}
=\frac{1}{2}\left(\mathcal{L}_{M\rightarrow G}+\mathcal{L}_{G\rightarrow M}\right),
\label{eq:lcon}
\end{equation}
with
\begin{equation}
\mathcal{L}_{M\rightarrow G}
=
-\frac{1}{B}\sum_{i=1}^{B}
\log
\frac{
\exp\!\left(\mathrm{sim}(i,i)/\tau\right)
}{
\sum_{j=1}^{B}\exp\!\left(\mathrm{sim}(i,j)/\tau\right)
},
\label{eq:lmg}
\end{equation}
where $\mathrm{sim}(i,j)$ denotes the cosine similarity between the projected image-side embedding of spot $i$ and the projected gene-side embedding of spot $j$, and $\tau$ is a learnable temperature parameter. The term $\mathcal{L}_{G\rightarrow M}$ is defined analogously by swapping the two modalities.

To preserve coordinate-induced tissue organization, CHRep further regularizes the gene-side representation with a topology-consistency term. For each mini-batch, a $k$NN graph is first constructed from the spot coordinates, and multi-hop neighborhood matrices are derived to encode local-to-global spatial structure \parencite{CHEN2024106287}. Let $\mathbf{A}^{(h)}\in\mathbb{R}^{B\times B}$ denote the adjacency matrix of the $h$-hop neighborhood in the current mini-batch. The coordinate-induced topology prior is defined as
\begin{equation}
\mathbf{A}_{\mathrm{topo}}
=
\sum_{h=1}^{H_{\mathrm{hop}}}\alpha_h \mathbf{A}^{(h)},
\label{eq:topo_prior}
\end{equation}
where $\alpha_h$ controls the contribution of different hop orders.

Let $\mathbf{F}_G$ denote the batch-wise gene-feature matrix formed by stacking $\{F_G(i)\}_{i=1}^{B}$. Its feature-similarity matrix is computed as
\begin{equation}
\mathbf{S}(\mathbf{F}_G)=\mathbf{F}_G\mathbf{F}_G^\top.
\label{eq:fg_similarity}
\end{equation}
The topology loss then encourages the similarity pattern induced by $\mathbf{F}_G$ to match the coordinate-induced topology prior:
\begin{equation}
\mathcal{L}_{\mathrm{spa}}
=
\left\|
\widetilde{\mathbf{S}}(\mathbf{F}_G)-\widetilde{\mathbf{A}}_{\mathrm{topo}}
\right\|_F^2,
\label{eq:lspa}
\end{equation}
where both matrices are normalized before comparison. In this way, the learned gene representation is encouraged to preserve higher-order tissue organization, and this structural bias is further conveyed to the morphology side through the contrastive alignment term in Eq.~\eqref{eq:lcon}.

Finally, the overall training objective combines regression, cross-modal alignment, and topology preservation:
\begin{equation}
\mathcal{L}
=
\lambda_{\mathrm{con}}\mathcal{L}_{\mathrm{con}}
+
\lambda_{\mathrm{reg}}\mathcal{L}_{\mathrm{reg}}
+
\lambda_{\mathrm{spa}}\mathcal{L}_{\mathrm{spa}}.
\label{eq:total_loss}
\end{equation}

\subsection{Post-hoc Calibration via an Estimate Module and a Correction Module}
\label{sec:posthoc}

Post-hoc calibration uses the frozen histology feature $\mathbf{z}_i=F_H(i)$ as the shared input to the Estimate Module and the Correction Module. After Stage I, the regression head $h_{\phi}$ is replaced by this calibration scheme. The Estimate Module retrieves expression profiles from the training gallery, and Stage II fits the Correction Module to learn an additive adjustment to the retrieved estimate. The coordinate and gene branches are used during Stage I.

Let $\mathcal{D}_{\mathrm{tr}}$ denote the training set for a given fold. The training gallery contains frozen histology features and their standardized expression vectors. For a query spot $i$, cosine similarity to a gallery spot $j\in\mathcal{D}_{\mathrm{tr}}$ is computed as
\begin{equation}
s_{ij}
=
\frac{\mathbf{z}_i^\top\mathbf{z}_j}
{\|\mathbf{z}_i\|_2\,\|\mathbf{z}_j\|_2}.
\label{eq:teacher_sim}
\end{equation}
Let $\mathcal{N}_K(i)$ denote the indices of the top-$K$ gallery neighbors with the highest similarity scores. The Estimate Module forms a similarity-weighted non-parametric prediction:
\begin{equation}
\begin{aligned}
w_{ij} &= \frac{\exp(s_{ij}/\tau_t)}{\sum_{j'\in\mathcal{N}_K(i)}\exp(s_{ij'}/\tau_t)},\\
\hat{\mathbf{g}}^{(E)}_i &= \sum_{j\in\mathcal{N}_K(i)} w_{ij}\,\tilde{\mathbf{g}}_j,
\end{aligned}
\label{eq:teacher_pred}
\end{equation}
where $\tau_t$ controls the sharpness of neighbor aggregation. During calibrator fitting, queries are drawn from $\mathcal{D}_{\mathrm{tr}}$ and each query is excluded from its retrieval candidates:
\begin{equation}
\mathcal{N}_K(i)\subseteq \mathcal{D}_{\mathrm{tr}}\setminus\{i\}.
\label{eq:no_self}
\end{equation}

A lightweight Correction Module $r_{\eta}$ refines the retrieval-based estimate by predicting an additive residual from the frozen query representation:
\begin{equation}
\Delta_i=r_{\eta}(\mathbf{z}_i), \qquad
\hat{\mathbf{g}}_i=\hat{\mathbf{g}}^{(E)}_i+\Delta_i,
\label{eq:correction_form}
\end{equation}
where $r_{\eta}$ is implemented as a small MLP. The residual magnitude is regularized as
\begin{equation}
\mathcal{L}_{\Delta}
=
\frac{1}{B}\sum_{i=1}^{B}\left\|\Delta_i\right\|_2^2.
\label{eq:delta_reg}
\end{equation}

The correction module is optimized while keeping the image branch frozen. Its objective applies the same MSE$+$MAE$+$PCC form as Eq.~\eqref{eq:lreg} to the calibrated predictions, together with residual regularization:
\begin{equation}
\mathcal{L}_{\mathrm{cal}}
=
\mathcal{L}_{\mathrm{reg}}^{\mathrm{cal}}
+
\lambda_{\Delta}\mathcal{L}_{\Delta},
\label{eq:lcal}
\end{equation}
where $\mathcal{L}_{\mathrm{reg}}^{\mathrm{cal}}$ reuses the same correlation-aware regression form as Eq.~\eqref{eq:lreg}, but is computed on the calibrated predictions in Eq.~\eqref{eq:correction_form}. Only the parameters $\eta$ of the correction module are updated in this stage.

At inference, the frozen image encoder maps each query patch from the held-out slide to $\mathbf{z}_i$. The Estimate Module retrieves neighbors from the training gallery and computes $\hat{\mathbf{g}}_i^{(E)}$ using Eq.~\eqref{eq:teacher_pred}; the fixed Correction Module supplies the residual in Eq.~\eqref{eq:correction_form}. Gallery construction and calibration training use the training slides of each fold.

\section{Experiments and Results}

\subsection{Datasets}
Experiments are conducted on three paired histology--spatial transcriptomics cohorts spanning distinct tissues and acquisition settings, covering both tumor microenvironment and breast cancer scenarios. Specifically, the evaluation is performed on the HER2+ breast cancer dataset with 32 tissue sections, the cSCC dataset with 12 tissue sections, and the Alex+10x dataset with 9 slices. Each cohort provides, for every capture location (spot), an H\&E-derived image patch centered at the spot, the corresponding spot-level gene expression vector, and the 2D spatial coordinates. A leave-one-slide-out protocol is adopted for evaluation, and performance on the held-out slide is reported using both correlation- and error-based metrics.

cSCC \parencite{ji2020multimodal} comprises spatially resolved transcriptomes measured on cutaneous squamous cell carcinoma specimens with matched histology. The cohort contains multiple patients with replicate tissue sections, providing a clinically relevant setting in which morphology varies across sections and patients and cross-slide generalization is essential. The standard preprocessing pipeline used in prior histology-to-transcriptome prediction studies is followed to obtain spot-level normalized expression targets and aligned H\&E patches.

HER2+ \parencite{andersson2020spatial} is a breast cancer spatial transcriptomics cohort with matched histology and spot coordinates and is widely used as a benchmark for histology-to-expression prediction. The commonly adopted subset of tissue sections with consistent spot layouts and reliable image--expression pairing is used, and PCC-based metrics are reported for both all genes and highly expressed genes.

Alex+10x \parencite{wu2021single} is a 10x Visium cohort with paired H\&E images and spot-level expression profiles. Leave-one-slide-out evaluation is performed across all available slides using the same gene subset and normalization convention throughout.

\subsection{Compared Methods}
Comparisons are conducted against representative histology-to-expression predictors that reflect the main modeling trends in this area. Regression-based baselines include ST-Net \parencite{he2020integrating}, whereas contextual modeling is represented by HisToGene \parencite{pang2021leveraging}. To account for spatial interactions, Hist2ST \parencite{zeng2022spatial} and hybrid architectures such as THItoGene \parencite{jia2023thitogene} are included. Cross-modal alignment approaches that learn shared image--expression embedding spaces are also considered, including retrieval-based BLEEP \parencite{xie2023spatially} and contrastive learning-based mclSTExp \parencite{min2024multimodal}. Two additional baselines are included on all three datasets: DFCE-KanT, which combines feature-channel enhancement with KAN-based spatial and cross-modal modeling \parencite{li2026dfce}, and SciSt, which incorporates single-cell-reference-derived expression information alongside histopathological features \parencite{li2025scist}. DFCE-KanT and SciSt are reproduced in this study on all three datasets, and their reported results are averaged over the same three random seeds. For each method, the authors' recommended settings are adopted when available; otherwise, hyperparameters are tuned using training/validation splits only.
\begin{table*}[t]

\caption{Mean Pearson correlation coefficients (PCCs) on all considered genes (ACG) and the top-50 highly expressed genes (HEG@50), along with mean squared error (MSE) and mean absolute error (MAE), across the HER2+, cSCC, and Alex+10x datasets.}

\label{tab:main_three_datasets}
\small
\setlength{\tabcolsep}{4pt}
\renewcommand{\arraystretch}{1.12}

\begin{tabular*}{\textwidth}{@{\extracolsep{\fill}}>{\raggedright\arraybackslash}p{0.30\textwidth}cccc@{}}
\toprule

\multicolumn{1}{c}{\multirow{2}{*}{Methods}}& \multicolumn{4}{c}{HER2+} \\
\cmidrule(lr){2-5}
 & \shortstack{PCC\\(ACG)$\uparrow$} & \shortstack{PCC\\(HEG@50)$\uparrow$} & MSE$\downarrow$ & MAE$\downarrow$ \\
\midrule
ST-Net~\parencite{he2020integrating}     & $0.0561$ & $0.0134$ & $0.5312$ & $0.6306$ \\
HisToGene~\parencite{pang2021leveraging}   & $0.0842$ & $0.0711$ & $0.5202$ & $0.6422$ \\
Hist2ST \parencite{zeng2022spatial}     & $0.1443$ & $0.1849$ & $0.5135$ & $0.6087$ \\
THItoGene~\parencite{jia2023thitogene}   & $0.1726$ & $0.2809$ & $0.5012$ & $0.5956$ \\
BLEEP~\parencite{xie2023spatially}      & $0.1873$ & $0.2909$ & $0.6015$ & $0.5824$ \\
mclSTExp~\parencite{min2024multimodal} & $0.2259$ & $0.3704$ & $0.5964$ & $0.5876$ \\
DFCE-KanT~\parencite{li2026dfce}
            & $0.2082$ & $0.3401$ & $0.6296$ & $0.6035$ \\
SciSt~\parencite{li2025scist}
            & $0.1658$ & $0.2215$ & {\boldmath$0.2602$} & {\boldmath$0.3740$} \\
\textbf{CHRep (Ours)}
            & {\boldmath$0.2733$} & {\boldmath$0.4634$} & $0.5616$ & $0.5002$ \\

\midrule

\multicolumn{1}{c}{\multirow{2}{*}{Methods}} & \multicolumn{4}{c}{cSCC} \\
\cmidrule(lr){2-5}
 & \shortstack{PCC\\(ACG)$\uparrow$} & \shortstack{PCC\\(HEG@50)$\uparrow$} & MSE$\downarrow$ & MAE$\downarrow$ \\
\midrule
ST-Net~\parencite{he2020integrating}      & $0.0012$ & $0.0018$ & $0.6806$ & $0.6404$ \\
HisToGene~\parencite{pang2021leveraging}   & $0.0771$ & $0.0919$ & $0.6805$ & $0.6234$ \\
Hist2ST~\parencite{zeng2022spatial}     & $0.1838$ & $0.2175$ & $0.6748$ & $0.6107$ \\
THItoGene~\parencite{jia2023thitogene}   & $0.2373$ & $0.2719$ & $0.6546$ & $0.6012$ \\
BLEEP~\parencite{xie2023spatially}       & $0.2449$ & $0.3122$ & $0.5163$ & $0.5399$ \\
mclSTExp~\parencite{min2024multimodal} & $0.3228$ & $0.4247$ & $0.4310$ & $0.5101$ \\
DFCE-KanT~\parencite{li2026dfce} & $0.1083$ & $0.1571$ & $0.4028$ & $0.5012$ \\
SciSt~\parencite{li2025scist} & $0.2742$ & $0.2595$ & $0.4350$ & $0.4941$ \\
\textbf{CHRep (Ours)}
            & {\boldmath$0.3532$} & {\boldmath$0.4682$} & {\boldmath$0.3983$} & {\boldmath$0.4048$} \\

\midrule

\multicolumn{1}{c}{\multirow{2}{*}{Methods}}& \multicolumn{4}{c}{Alex+10x} \\
\cmidrule(lr){2-5}
 & \shortstack{PCC\\(ACG)$\uparrow$} & \shortstack{PCC\\(HEG@50)$\uparrow$} & MSE$\downarrow$ & MAE$\downarrow$ \\
\midrule
ST-Net~\parencite{he2020integrating}      & $0.0009$ & $0.0452$ & $0.4721$ & $0.5042$ \\
HisToGene~\parencite{pang2021leveraging}   & $0.0618$ & $0.0984$ & $0.4565$ & $0.4973$ \\
Hist2ST~\parencite{zeng2022spatial}      & $0.1299$ & $0.1784$ & $0.3788$ & $0.4492$ \\
THItoGene~\parencite{jia2023thitogene}   & $0.1384$ & $0.2156$ & $0.3672$ & $0.4315$ \\
BLEEP~\parencite{xie2023spatially}        & $0.1552$ & $0.2825$ & $0.2593$ & $0.4050$ \\
mclSTExp~\parencite{min2024multimodal} & $0.1882$ & $0.3245$ & $0.2452$ & $0.4018$ \\
DFCE-KanT~\parencite{li2026dfce} & $0.1063$ & $0.1873$ & $0.2793$ & $0.4324$ \\
SciSt~\parencite{li2025scist} & $0.0703$ & $0.1437$ & $0.3311$ & $0.4162$ \\
\textbf{CHRep (Ours)}
            & {\boldmath$0.2718$} & {\boldmath$0.4659$} & {\boldmath$0.2102$} & {\boldmath$0.3545$} \\

\bottomrule
\end{tabular*}

\par\vspace{3pt}
{\footnotesize\raggedright
\textit{Notes.} CHRep and the DFCE-KanT and SciSt baselines reproduced in this study are evaluated using the same three random seeds, and their results are reported as means. Bold indicates the best displayed mean for each dataset and metric.
\par}
\end{table*}

\begin{figure*}[!t]
    \centering
    \includegraphics[width=\textwidth]{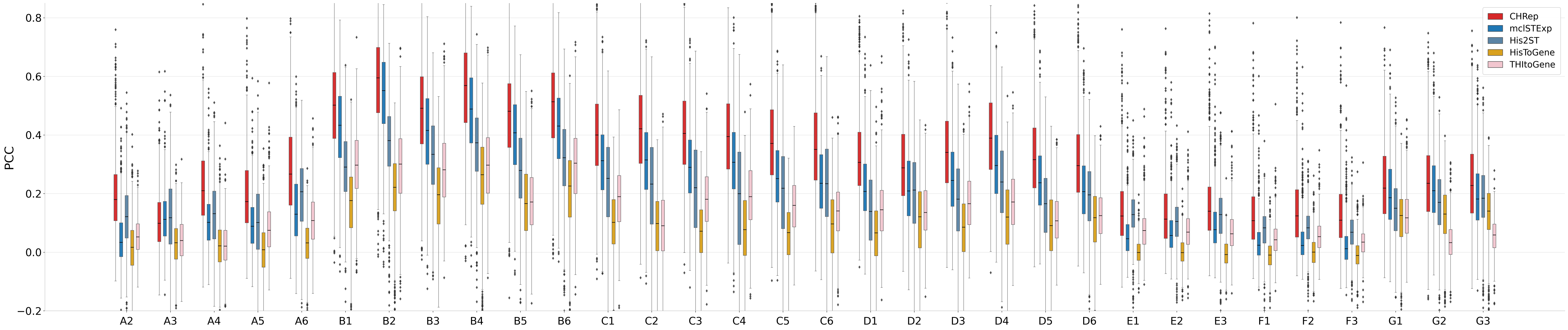}
    \caption{Per-slide distribution of gene-wise Pearson correlations on HER2+ (785 genes). 
   For each slide, boxplots of gene-wise PCC values between predicted and ground-truth expression are reported for CHRep and representative baselines.
    The distributions summarize gene-wise prediction accuracy within each held-out slide.}
    \label{fig:pcc_boxplot_slide}
\end{figure*}
\begin{figure*}[!t]
    \centering
    \includegraphics[width=\textwidth]{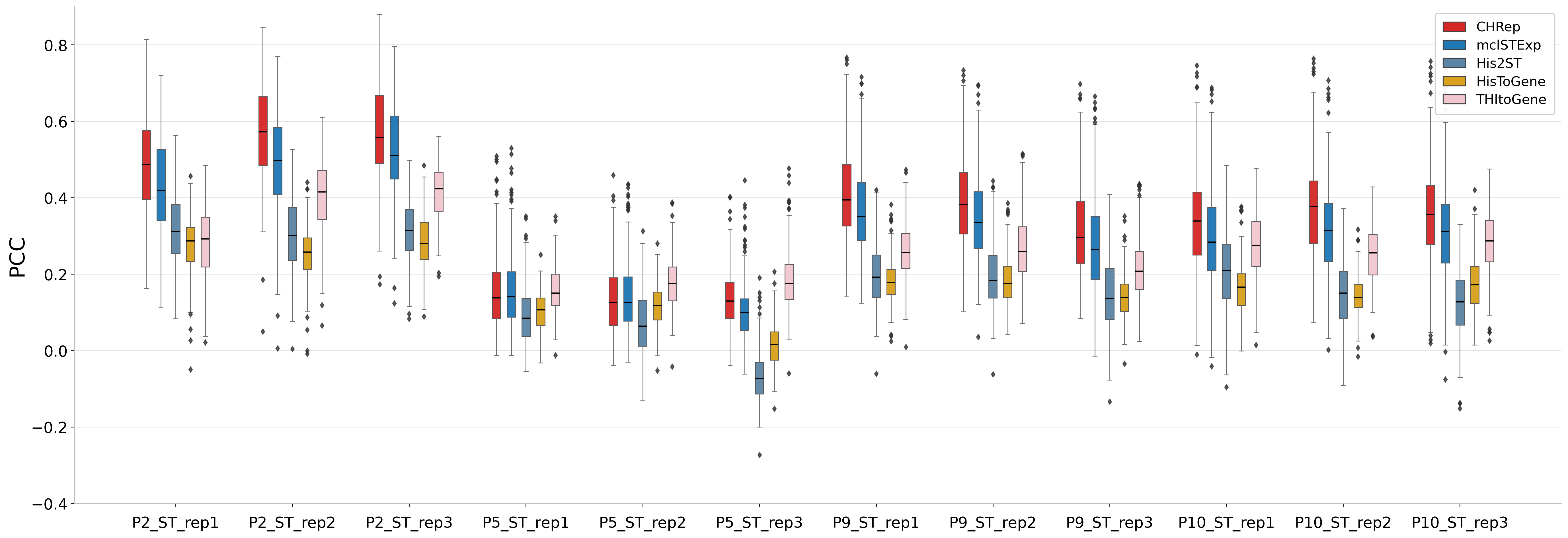}
    \caption{Per-slide distribution of gene-wise PCC on cSCC under leave-one-slide-out evaluation.}
    \label{fig:cscc_box}
\end{figure*}

\subsection{Evaluation Metrics}

Both correlation-based and error-based metrics are reported under the same convention as that used in training. Let $\mathbf{g}_i\in\mathbb{R}^{G}$ denote the log-normalized ground-truth expression for spot $i$, and let
$\tilde{\mathbf{g}}_i=(\mathbf{g}_i-\boldsymbol{\mu})\oslash(\boldsymbol{\sigma}+\epsilon)$ denote the per-gene standardized expression, where $(\boldsymbol{\mu},\boldsymbol{\sigma})$ are computed from the training slides of the current fold. Let $\hat{\mathbf{g}}_i\in\mathbb{R}^{G}$ denote the predicted standardized expression for spot $i$. All metrics below are computed in the standardized space $\tilde{\mathbf{g}}$ to maintain a consistent per-gene scale across slides \parencite{wang2025benchmarking}. The Pearson correlation coefficient (PCC) is invariant to positive affine rescaling, whereas the mean squared error (MSE) and mean absolute error (MAE) depend on scale.

\textbf{Gene-wise Pearson correlation.}
For each gene $j$, the Pearson correlation coefficient across the $N$ spots of a held-out slide is computed as
\begin{equation}
\mathrm{PCC}_j
=
\frac{\sum_{i=1}^{N}\left(\tilde{g}_{i,j}-\overline{\tilde{g}}_{\cdot,j}\right)\left(\hat{g}_{i,j}-\bar{\hat{g}}_{\cdot,j}\right)}
{\sqrt{\sum_{i=1}^{N}\left(\tilde{g}_{i,j}-\overline{\tilde{g}}_{\cdot,j}\right)^2}\;
 \sqrt{\sum_{i=1}^{N}\left(\hat{g}_{i,j}-\bar{\hat{g}}_{\cdot,j}\right)^2}},
\label{eq:pcc_gene}
\end{equation}
where $\overline{\tilde{g}}_{\cdot,j}=\frac{1}{N}\sum_{i=1}^{N}\tilde{g}_{i,j}$ and $\bar{\hat{g}}_{\cdot,j}=\frac{1}{N}\sum_{i=1}^{N}\hat{g}_{i,j}$.

For a gene subset $\mathcal{S}$, the set-level correlation metric is defined as the average of gene-wise Pearson correlation coefficients:
\begin{equation}
\mathrm{PCC}(\mathcal{S})
=
\frac{1}{|\mathcal{S}|}\sum_{j\in\mathcal{S}}\mathrm{PCC}_j.
\label{eq:pcc_set}
\end{equation}
In particular, ACG denotes all considered genes, i.e., the full HVG set used in evaluation. Therefore, $\mathrm{PCC}(\mathrm{ACG})$ represents the average gene-wise Pearson correlation coefficient over the full evaluation gene set.

\textbf{Highly expressed genes (HEG@K).}
For each held-out slide, HEG@K denotes the top-$K$ genes ranked by their mean log-normalized ground-truth expression:
\begin{equation}
m_j=\frac{1}{N}\sum_{i=1}^{N}g_{i,j},\qquad
\mathcal{S}_{\mathrm{HEG@K}}=\mathrm{TopK}\left(\{m_j\}_{j=1}^{G}\right),
\label{eq:heg_set}
\end{equation}
The corresponding correlation metric is
\begin{equation}
\mathrm{PCC}(\mathrm{HEG@K})=\mathrm{PCC}\left(\mathcal{S}_{\mathrm{HEG@K}}\right).
\label{eq:pcc_heg}
\end{equation}

\textbf{Error metrics.}
Mean squared error (MSE) and mean absolute error (MAE) are also reported:
\begin{equation}
\mathrm{MSE}
=
\frac{1}{NG}\sum_{i=1}^{N}\left\|\hat{\mathbf{g}}_i-\tilde{\mathbf{g}}_i\right\|_2^2,
\qquad
\mathrm{MAE}
=
\frac{1}{NG}\sum_{i=1}^{N}\left\|\hat{\mathbf{g}}_i-\tilde{\mathbf{g}}_i\right\|_1.
\label{eq:mse_mae}
\end{equation}
Correlation metrics reflect gene-wise trend consistency, whereas MSE/MAE quantify absolute deviations; reporting both provides a balanced assessment under noisy and sparse ST measurements.

\subsection{Implementation Details}
Results for CHRep and the DFCE-KanT and SciSt baselines reproduced in this study, as well as all subsequent ablation and sensitivity results, are reported as means over the same three random seeds. Metrics are first averaged equally over held-out slides within each seed and then averaged over the three seeds. The fixed-feature calibration and sensitivity experiments vary Stage-II initialization while keeping the Stage-I features fixed.

CHRep uses two training stages. Each spot is represented by an H\&E-centered patch $\mathbf{x}_i\in\mathbb{R}^{3\times224\times224}$. During Stage I, the image encoder extracts $F_H(i)=f_{\theta}(\mathbf{x}_i)$, and the regression head predicts $\hat{\mathbf{g}}_i=h_{\phi}(F_H(i))\in\mathbb{R}^{G}$. Image augmentation includes random flips, rotations, and color jitter. A position-embedding branch encodes spatial coordinates and fuses them with histology features for cross-modal alignment. Coordinate-based $k$NN graphs and multi-hop neighborhoods define the topology target for gene features. Lightweight projection heads produce normalized image-side and gene-side embeddings for the symmetric contrastive objective.

During Stage II, the image encoder is frozen and training-spot embeddings form the gallery bank. The Estimate Module aggregates the standardized expression vectors of the top-$K$ neighbors to obtain $\hat{\mathbf{g}}_i^{(E)}$. The Correction Module is trained on the training slides to predict $\Delta_i=r_{\eta}(\mathbf{z}_i)$, yielding $\hat{\mathbf{g}}_i=\hat{\mathbf{g}}_i^{(E)}+\Delta_i$. Its objective combines correlation-aware regression with residual-magnitude regularization. Hyperparameters are selected using the training/validation portion of each fold.

For the fixed-feature calibration experiments, the direct and residual predictors use two 512-unit hidden layers with GELU activations and dropout of 0.1, with LayerNorm applied after the first activation. Both predictors are trained for 90 epochs using Adam with a learning rate of $10^{-4}$, weight decay of $10^{-4}$, and a batch size of 512. The PCC-loss coefficient is set to 0.3. Retrieval uses $L_2$-normalized embeddings, $K=200$ unless otherwise specified, and a softmax temperature of $\tau_t=1$. Training queries exclude themselves from the gallery. A separate slide is reserved for validation within each fold, and the checkpoint with the highest validation PCC(ACG) is selected.

\begin{figure}[t]
    \centering
    \includegraphics[width=\columnwidth]{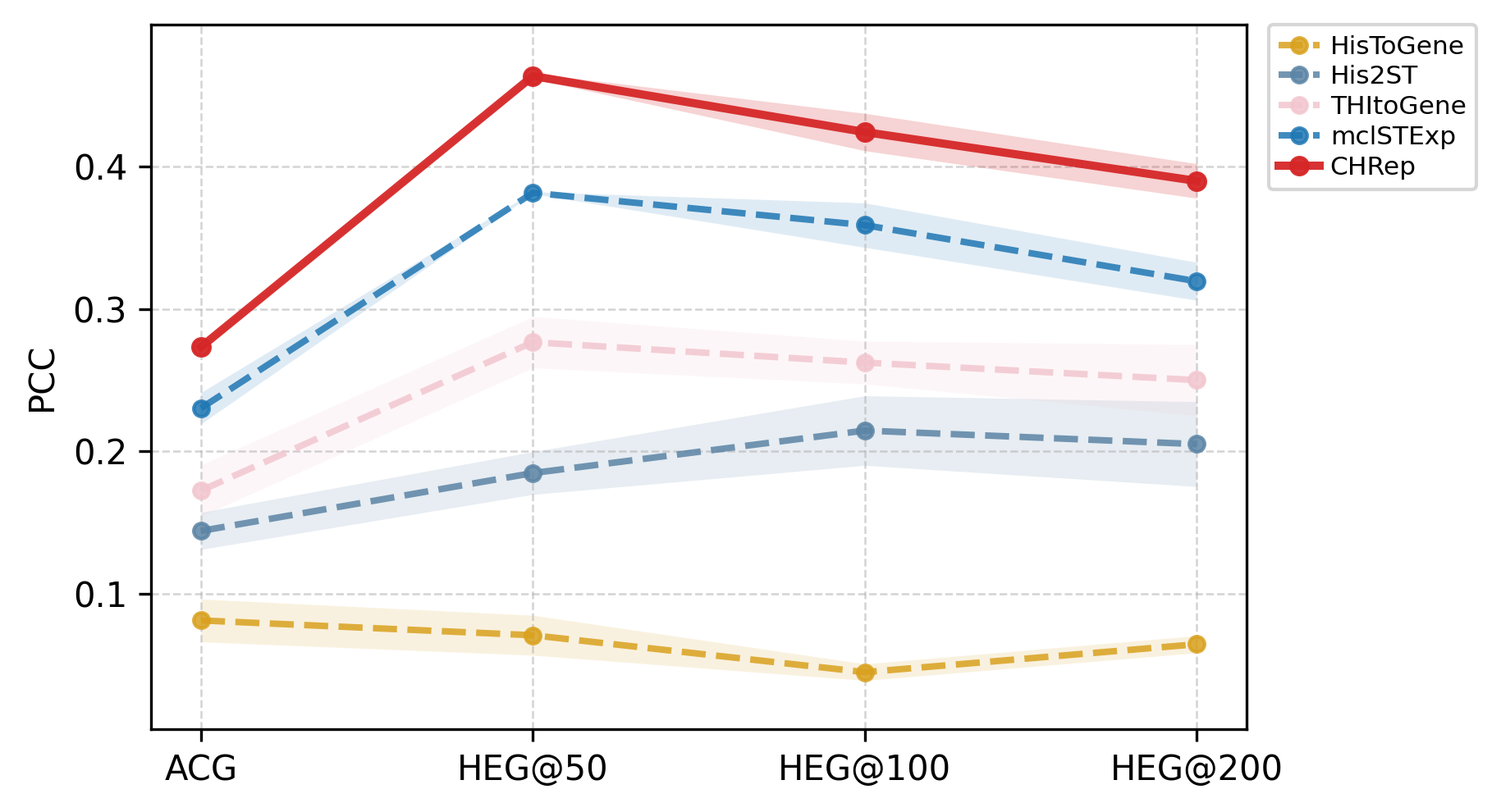}
    \caption{HER2+ performance as a function of the gene set. PCC on ACG and HEG@K with $K\in\{50,100,200\}$ is reported. Shaded bands indicate the standard deviation across folds.}
    \label{fig:her2_heg_curve}
\end{figure}

\subsection{Comparative Experiments}
\label{sec:main_results}

Table~\ref{tab:main_three_datasets} reports leave-one-slide-out performance on HER2+, cSCC, and Alex+10x using correlation-based and error-based metrics, with DFCE-KanT and SciSt included on all three datasets. CHRep achieves the highest PCC(ACG) and PCC(HEG@50) among the reported methods across the three cohorts. It also achieves the lowest MSE and MAE on both cSCC and Alex+10x, demonstrating improvements in both gene-wise correlation and absolute prediction accuracy on these cohorts.

On HER2+, CHRep achieves PCC(ACG) and PCC(HEG@50) values of 0.2733 and 0.4634, respectively, compared with 0.2082 and 0.3401 for DFCE-KanT and 0.1658 and 0.2215 for SciSt. CHRep obtains an MSE of 0.5616 and an MAE of 0.5002, while SciSt achieves the lowest corresponding errors of 0.2602 and 0.3740.

Fig.~\ref{fig:pcc_boxplot_slide} shows the per-slide distribution of gene-wise PCCs across 785 genes on HER2+. CHRep exhibits higher medians and reduced low-correlation tails on most slides, showing that its correlation gains extend across multiple held-out sections.

Fig.~\ref{fig:her2_heg_curve} further evaluates highly expressed genes on HER2+ using $\mathrm{PCC}(\mathrm{HEG@K})$ with $K\in\{50,100,200\}$. CHRep achieves the highest correlation among the displayed methods at each evaluated gene-set size, showing that the improvement extends from HEG@50 to HEG@200.

\begin{figure*}[!t]
    \centering
    \includegraphics[width=0.7\textwidth]{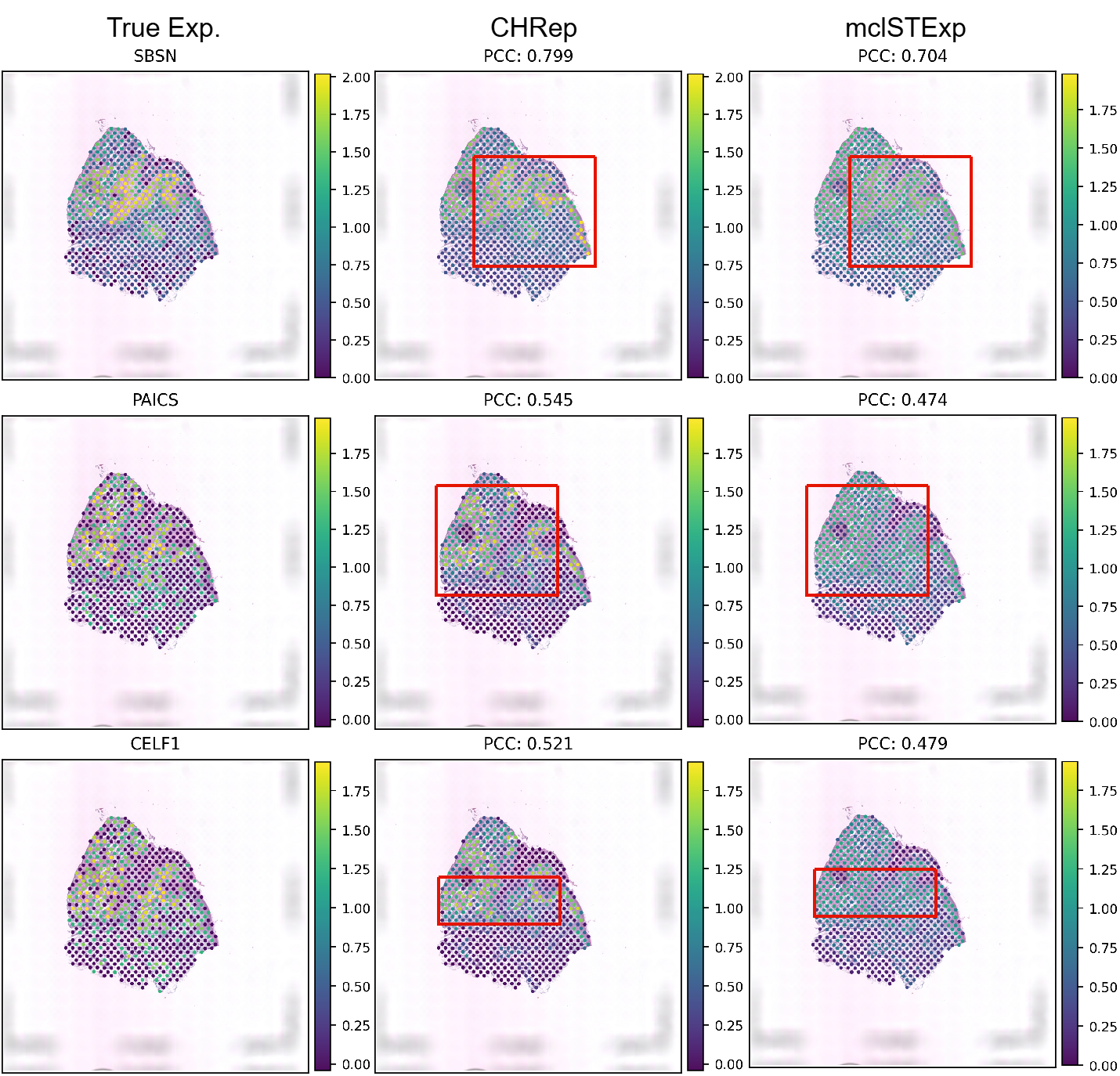}
\caption{Qualitative comparison of predicted spatial expression patterns on a held-out cSCC slide. Rows correspond to the genes \textit{SBSN}, \textit{PAICS}, and \textit{CELF1}, while columns show the ground-truth expression, CHRep prediction, and mclSTExp prediction, respectively, with spot-level expression values overlaid on the same H\&E image. The PCC values reported above the predicted maps denote the gene-wise Pearson correlation coefficients on the displayed slide. Red boxes highlight representative local regions for cross-method visual comparison. Compared with mclSTExp, CHRep produces spatial patterns that more closely match the ground truth, with clearer recovery of localized high-expression regions and better preservation of spatially coherent expression variation.}
    \label{fig:cscc_fixed_slide_qual}
\end{figure*}

On cSCC, CHRep achieves the highest PCC(ACG) and PCC(HEG@50), reaching 0.3532 and 0.4682, respectively. mclSTExp is the strongest correlation baseline, with corresponding values of 0.3228 and 0.4247, whereas SciSt obtains 0.2742 and 0.2595 and DFCE-KanT obtains 0.1083 and 0.1571. CHRep achieves the lowest MSE of 0.3983, followed by DFCE-KanT at 0.4028, and the lowest MAE of 0.4048. Thus, CHRep achieves the best results across all four metrics on cSCC among the compared methods.

Fig.~\ref{fig:cscc_box} shows higher median gene-wise PCCs for CHRep on most held-out cSCC slides. The spatial expression maps in Fig.~\ref{fig:cscc_fixed_slide_qual} further illustrate its recovery of localized enrichment and tissue-associated expression patterns.

On Alex+10x, CHRep achieves the best results across all four metrics, with PCC(ACG), PCC(HEG@50), MSE, and MAE values of 0.2718, 0.4659, 0.2102, and 0.3545, respectively. The strongest baseline across these metrics is mclSTExp, with corresponding values of 0.1882, 0.3245, 0.2452, and 0.4018. Relative to mclSTExp, CHRep improves PCC(ACG) and PCC(HEG@50) by 44.4\% and 43.6\%, respectively, while reducing MSE by 14.3\% and MAE by 11.8\%. Taken together, the results show consistent correlation improvements across cohorts and simultaneous gains in correlation and absolute error on cSCC and Alex+10x.

\begin{table*}[t]
\centering
\caption{Comparison of post-hoc calibration variants across three datasets.}
\label{tab:stage2_ablation}
\small
\setlength{\tabcolsep}{4pt}
\renewcommand{\arraystretch}{1.12}

\begin{tabular*}{\textwidth}{
@{\extracolsep{\fill}}
>{\raggedright\arraybackslash}p{0.30\textwidth}
cccc@{}
}
\toprule
\multicolumn{5}{c}{HER2+} \\
\cmidrule(lr){1-5}
Variants
& \shortstack{PCC\\(ACG)$\uparrow$}
& \shortstack{PCC\\(HEG@50)$\uparrow$}
& MSE$\downarrow$
& MAE$\downarrow$ \\
\midrule
Estimate only
& 0.2460 & 0.4124 & 0.5893 & 0.5767 \\
Direct MLP
& 0.2542 & 0.4251 & 0.6115 & 0.5018 \\
Estimate + Correction (w/o reg.)
& 0.2535 & 0.4329 & 0.6073 & 0.5068 \\
\textbf{CHRep (Ours)}
& \textbf{0.2733} & \textbf{0.4634}
& \textbf{0.5616} & \textbf{0.5002} \\
\midrule

\multicolumn{5}{c}{cSCC} \\
\cmidrule(lr){1-5}
Variants
& \shortstack{PCC\\(ACG)$\uparrow$}
& \shortstack{PCC\\(HEG@50)$\uparrow$}
& MSE$\downarrow$
& MAE$\downarrow$ \\
\midrule
Estimate only
& 0.3440 & 0.4410 & 0.4089 & 0.4704 \\
Direct MLP
& 0.3432 & 0.4654 & 0.4250 & \textbf{0.3970} \\
Estimate + Correction (w/o reg.)
& 0.3403 & 0.4520 & 0.4225 & 0.4030 \\
\textbf{CHRep (Ours)}
& \textbf{0.3532} & \textbf{0.4682}
& \textbf{0.3983} & 0.4048 \\
\midrule

\multicolumn{5}{c}{Alex+10x} \\
\cmidrule(lr){1-5}
Variants
& \shortstack{PCC\\(ACG)$\uparrow$}
& \shortstack{PCC\\(HEG@50)$\uparrow$}
& MSE$\downarrow$
& MAE$\downarrow$ \\
\midrule
Estimate only
& 0.2158 & 0.3551 & 0.2414 & 0.4071 \\
Direct MLP
& 0.2584 & 0.4220 & 0.2428 & 0.3575 \\
Estimate + Correction (w/o reg.)
& 0.2515 & 0.4127 & 0.2488 & 0.3740 \\
\textbf{CHRep (Ours)}
& \textbf{0.2718} & \textbf{0.4659}
& \textbf{0.2102} & \textbf{0.3545} \\
\bottomrule
\end{tabular*}

\par\vspace{3pt}
{\footnotesize\raggedright
\textit{Notes.} All variants are evaluated under the same experimental conditions using the same three random seeds, and the reported values are their means. Within each fold, all variants use identical fixed Stage-I features, data splits, and preprocessing; all other training settings are held constant except for the calibration component being compared.
Retrieval-based variants use $K=200$.
Bold indicates the best result within each dataset and metric.
\par}
\end{table*}
\subsection{Analysis of Key Design Components}
\label{sec:ablation}

The following analyses evaluate post-hoc calibration on all three datasets (Table~\ref{tab:stage2_ablation}), the representation objective on cSCC (Table~\ref{tab:ablate_repr_cscc}), and retrieval-neighborhood sensitivity on cSCC (Table~\ref{tab:sensitivity_k}). Results are averaged over the same three random seeds. Calibration variants share identical fixed Stage-I features within each fold.

\subsection{Effect of the Post-hoc Calibration Design}
\label{sec:ablate_cal}

Table~\ref{tab:stage2_ablation} compares four calibration variants across the three datasets. CHRep achieves the best results across all four metrics on HER2+ and Alex+10x. On cSCC, it obtains the highest PCC(ACG) and PCC(HEG@50) of 0.3532 and 0.4682, respectively, together with the lowest MSE of 0.3983. Direct MLP achieves the lowest MAE of 0.3970 on this dataset, compared with 0.4048 for CHRep.

Relative to Estimate only, CHRep improves both correlation metrics and reduces MSE and MAE across all three datasets, demonstrating the effectiveness of the complete calibration design. Adding magnitude regularization to Estimate + Correction further improves PCC(ACG), PCC(HEG@50), and MSE on every dataset. On HER2+, for example, regularization increases PCC(ACG) from 0.2535 to 0.2733 and PCC(HEG@50) from 0.4329 to 0.4634, while reducing MSE from 0.6073 to 0.5616. These results support combining retrieval-based estimation with regularized residual correction.

\subsection{Contribution of the Representation Objective}
\label{sec:ablate_repr}

\begin{table*}[t]
\caption{Contribution analysis of alternative representation-objective configurations on cSCC. ``Topology'' denotes the coordinate-induced (multi-hop) topology preservation term; ``Contrastive'' denotes the symmetric cross-modal alignment objective.}
\centering
\small
\setlength{\tabcolsep}{3pt}
\renewcommand{\arraystretch}{1.08}
\begin{tabular*}{\textwidth}{@{\extracolsep{\fill}}lccccccc@{}}
\toprule
\multirow{2}{*}{Variants} & \multicolumn{3}{c}{Representation Components} & \multicolumn{4}{c}{cSCC} \\
\cmidrule(lr){2-4}\cmidrule(lr){5-8}
 & Topology & Regression & Contrastive & \shortstack{PCC\\(ACG)$\uparrow$} &\shortstack{PCC\\(HEG@50)$\uparrow$} & MSE$\downarrow$ & MAE$\downarrow$ \\
\midrule
Regression + Contrastive            & $\xmark$ & $\cmark$ & $\cmark$ & 0.3412 & 0.4482 & 0.4303 & 0.4272 \\
Topology + Regression               & $\cmark$ & $\cmark$ & $\xmark$ & 0.3321 & 0.4272 & 0.4292 & 0.4103 \\
Contrastive + Topology  & $\cmark$ & $\xmark$ & $\cmark$ & 0.3482 & 0.4563 & 0.4250 & 0.4115 \\
\textbf{Full objective (Ours)}      & $\cmark$ & $\cmark$ & $\cmark$ & \textbf{0.3522} & \textbf{0.4663} & \textbf{0.4191} & \textbf{0.4051} \\
\bottomrule
\end{tabular*}

\label{tab:ablate_repr_cscc}
\par\vspace{3pt}
{\footnotesize\raggedright
\textit{Notes.} Results are averaged over the same three random seeds in a separate representation-objective experiment. The full objective serves as the reference for its ablations. Bold indicates the best value in each metric.
\par}
\end{table*}

Table~\ref{tab:ablate_repr_cscc} shows that the full objective achieves the best performance across all four metrics on cSCC. Adding topology regularization to Regression+Contrastive improves PCC(ACG) from $0.3412$ to $0.3522$, together with higher PCC(HEG@50) and lower MSE and MAE. Contrastive alignment connects gene-side spatial supervision to the image representation, while regression provides direct expression supervision. Removing either component reduces performance, supporting the complementary contributions of the three objectives.

\subsection{Effect of Retrieval Neighborhood Size on CHRep}
\label{sec:sensitivity_k}

Table~\ref{tab:sensitivity_k} reports CHRep performance with different retrieval neighborhood sizes on cSCC. As $K$ increases from 10 to 200, PCC(ACG) improves from 0.3170 to 0.3532 and PCC(HEG@50) increases from 0.4276 to 0.4682, while MAE decreases from 0.4514 to 0.4048. The improvement in PCC(ACG) becomes smaller between $K=100$ and $K=200$, increasing from 0.3513 to 0.3532. Among the tested settings, $K=200$ achieves the best performance across all four metrics, including the lowest MSE of 0.3983, supporting its use as the default retrieval neighborhood size.

\begin{table}[H]
\centering
\caption{CHRep performance with different retrieval neighborhood sizes on cSCC.}
\label{tab:sensitivity_k}
\small
\setlength{\tabcolsep}{2.5pt}
\renewcommand{\arraystretch}{1.12}

\begin{tabular*}{\columnwidth}{@{\extracolsep{\fill}}ccccc@{}}
\toprule
$K$
& \shortstack{PCC\\(ACG)$\uparrow$}
& \shortstack{PCC\\(HEG@50)$\uparrow$}
& MSE$\downarrow$
& MAE$\downarrow$ \\
\midrule
10  & 0.3170 & 0.4276 & 0.4442 & 0.4514 \\
50  & 0.3413 & 0.4562 & 0.4215 & 0.4149 \\
100 & 0.3513 & 0.4613 & 0.4217 & 0.4066 \\
200 & \textbf{0.3532} & \textbf{0.4682}
    & \textbf{0.3983} & \textbf{0.4048} \\
\bottomrule
\end{tabular*}

\par\vspace{3pt}
{\footnotesize\raggedright
\textit{Notes.} All neighborhood-size settings are evaluated using the same three random seeds, and the reported values are their means.
Bold indicates the best result for each metric.
\par}
\end{table}

\section{Conclusion}
CHRep introduces a two-stage framework for spatial gene expression prediction from H\&E histology, combining structure-aware representation learning with post-hoc calibration. The representation objective integrates correlation-aware regression, cross-modal alignment, and coordinate-induced topology regularization. Similarity-guided estimation and regularized residual correction then refine predictions using frozen histology features. Under leave-one-slide-out evaluation on HER2+, cSCC, and Alex+10x, CHRep achieves the highest PCC(ACG) and PCC(HEG@50) among the compared methods, together with the lowest MSE and MAE on cSCC and Alex+10x. Fixed-feature comparisons show that the complete calibration design improves all four metrics over retrieval alone across the three datasets, while the representation-objective analysis demonstrates the complementary contributions of its three components. These findings support the integration of spatially informed cross-modal representation learning with lightweight prediction refinement for histology-based spatial gene expression prediction.

\section*{Funding}
This work was supported by the National Natural Science Foundation of China under Grant 62401069.


\FloatBarrier
\printbibliography[title={References}]
\end{document}